\documentclass[conference,a4paper]{IEEEtran}

\usepackage{cite}
\usepackage{xcolor}
\usepackage{epsfig}
\usepackage{amsmath}
\usepackage{amssymb}
\usepackage{graphicx}
\usepackage{textcomp}
\usepackage{amsfonts}
\usepackage{stfloats}
\usepackage{enumitem}
\usepackage{algorithmic}

\graphicspath{ {./figures/} } 

\begin{document}

	
\title{On the Minimal Recognizable Image Patch}

\author{\IEEEauthorblockN{1\textsuperscript{st} Mark Fonaryov}
\IEEEauthorblockA{\textit{Autonomous Systems and Robotics Program} \\
\textit{Technion}\\
Haifa, Israel \\
markf@campus.technion.ac.il}
\and
\IEEEauthorblockN{2\textsuperscript{nd} Michael Lindenbaum}
\IEEEauthorblockA{\textit{Department of Computer Science} \\
\textit{Technion}\\
Haifa, Israel \\
mic@cs.technion.ac.il}
}
\maketitle


\begin{abstract}
	In contrast to human vision, common recognition algorithms often fail on partially occluded images. We propose characterizing, empirically, the algorithmic limits by finding a minimal recognizable patch (MRP) that is by itself sufficient to recognize the image. A specialized deep network allows us to find the most informative patches of a given size, and serves as an experimental tool. A human vision study recently characterized related (but different) minimally recognizable configurations (MIRCs) \cite{ullman2016atoms}, for which we specify computational analogues (denoted cMIRCs). 
	The drop in human decision accuracy associated with size reduction of these MIRCs is substantial and  sharp. Interestingly, such sharp reductions were also found for the computational versions we specified. 
\end{abstract}


\section{Introduction} \label{sec:Introduction}
Deep neural networks (DNNs) provide the current state-of-the-art performance in many computer vision tasks, and especially in recognition  \cite{krizhevsky2012imagenet,girshick2014rich,he2016deep}.
In contrast to human recognition processes, which can successfully recognize small or only partially visible objects \cite{tang2018recurrent,ullman2016atoms}, the performance of neural networks quickly deteriorates when objects are partially occluded or cropped \cite{pepik2015holding,osherov2017increasing,geirhos2018generalisation}.

This raises a natural question: 
What is the minimal image part (or parts) that suffice for recognizing an object? 

In this work, we consider a special practical version of this question: what is the minimal size of a square sub-image (patch) that is sufficient for recognition using a convolutional neural network (CNN)?
We chose to consider a CNN as a substitute to general recognizability because, currently, CNN algorithms are  as good as if not better than any other algorithm at this task.
We actually examine that question for two types of patches:
\begin{description}
	\item [Globally minimal patch -] Here we look for a patch of minimal size that provides the correct (and best) categorization. We denote this sub-image the minimally recognizable patch, or MRP.
	\item [Locally minimal patch -] Here we look for a patch that suffices for correct categorization, but whose contained sub-patches, do not. This criterion follows the {\em minimally recognizable configuration} (MIRC) specified in \cite{ullman2016atoms} using human responses (see Sec. \ref{sec:Related Work}). We specify this patch computationally and correspondingly denote it cMIRC.
\end{description}

\begin{figure}[t]
	\centering
	\includegraphics[width = \columnwidth]{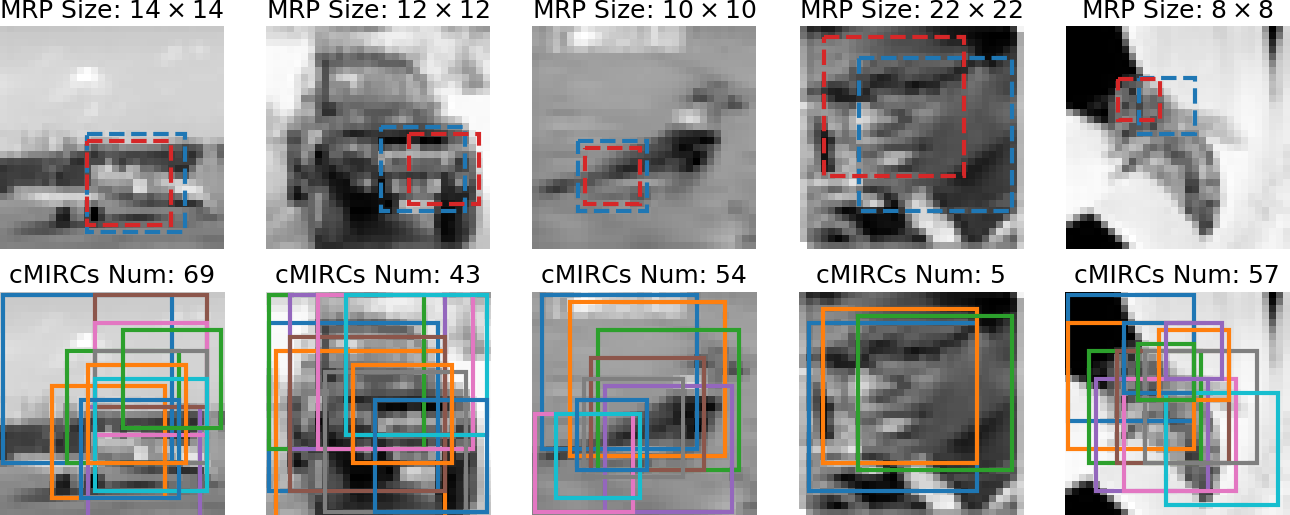}
	\caption{Examples of minimal patches. Top row: MRPs (blue) and best smaller unrecognizable patches (red). Bottom row: cMIRCs (excluding highly overlapping patches).}
	\label{fig:General Example}
\end{figure}

Note that cMIRCs are not unique and are also not of minimal size. 
The MRP on the other hand, is unique and is of minimal size, no larger than the size of the smallest cMIRC in the given image (see examples in Fig. \ref{fig:General Example}). Being minimal, it may be considered as one way for quantifying the minimal amount of information required for recognition. See Sec. \ref{sec:Related Work} for related studies. 

To find minimal recognizable patches, we designed a special neural architecture that identifies the most informative patch and classifies the image based on the information within it. Several variations of this patch-based classification (PBC) architecture, corresponding to different patch sizes and different ways of accumulating the local information, are considered. 
As expected, the minimal recognizable patches we found differ between and within categories, and increase in size for higher required accuracy.  

A particularly surprising finding of the MIRC paper  \cite{ullman2016atoms}, which motivated this study,  is that human recognition accuracy drops sharply and significantly with patch size, exactly for the size separating MIRCs and their sub-patches. This comes in contrast to algorithmic classifiers where the accuracy decreases smoothly with patch size, and suggests that a different mechanism is applied in the human system.  Interestingly, and in contrast to previous studies, we found similar sharp changes for MRPs and cMIRCs. 

This paper offers the following contributions:
\begin{itemize}
	\item PBC neural architecture that finds the most informative patch and uses it to categorize the entire image.  
	\item  Characterization of  globally and locally minimal patches sufficient for categorization.
	\item  An investigation of the confidence vs. patch size behaviour,  pointing at similarities between patch-based categorization in human vision and in algorithmic implementations. 
\end{itemize}

\section{Related Work} \label{sec:Related Work}
Local regions and features have been widely used by classical recognition algorithms and have provided improved immunity to pose change, partial occlusion, and in-class variance.
Prominent examples include image fragments \cite{vidal2003object}, visual bag of words models \cite{csurka2004visual} and the deformable-parts-models \cite{felzenszwalb2008discriminatively}.

Likewise, convolutional neural networks (CNNs) calculate local-scale descriptions, which are implicitly integrated into discriminative part representations and eventually globally combined into a decision \cite{krizhevsky2012imagenet,linsley2019learning,zeiler2014visualizing}.
The process carried out by CNNs of combining local features with increasingly more global and expressive structural information provides the current state-of-the-art performance in classification and detection of fully visible objects.
Yet, recognition accuracy of CNNs decreases when faced with partially visible objects, as happens for example in the presence of occlusions \cite{pepik2015holding,opitz2016grid,osherov2017increasing}, 

A CNN's robustness to partial occlusions can be improved by incorporating samples of occluded objects during training \cite{pepik2015holding}.
However, not every possible occlusion can be encountered in the training dataset as the occlusion distribution seems to follow a long tail \cite{wang2017fast}. In addition, the effects of such an approach are limited and do not generalize well \cite{geirhos2018generalisation}.
Specialized dedicated modifications to standard recognition architectures either exploit domain-specific prior knowledge \cite{zhang2018occlusion} or work by reducing the spatial support of their learned features \cite{osherov2017increasing}.
These approaches improve their partial object accuracy, which however is still considerably lower than that obtained with full-object visibility.

This deterioration is expected, yet it stands in contrast to human vision, in which object recognition is attained remarkably well, even when seeing only partial data.
First, low resolution is sufficient \cite{torralba200880} and  $32\times32$ color (or $64\times64$ gray-level) images are recognized well.
Second, human object recognition abilities remain robust when only small amounts of information are available due to heavy occlusion and $10\%$ visibility is adequate for performance well above chance \cite{tang2018recurrent}.

Recently, a psychophysical study \cite{ullman2016atoms} further showed that reliable human object recognition is possible even from small image-patches and specified a special class of minimal image patches.
These patches are minimal in the sense that they are recognizable, but their sub-patches, smaller by $20\%$, or identical patches with $20\%$ lower resolution, are not. That is, such patches, denoted  minimal recognizable configurations (MIRCs), are locally minimal. Interestingly, this study found that human recognition accuracy associated with the sub-patches was significantly lower than that associated with the MIRC itself. Tests with recognition algorithms, applied to MIRCs and to their sub-images, did not find a similar accuracy drop.
A follow-up on this work demonstrated that CNN classification for some patches, denoted fragile recognition images (FRIs), may be changed due to small translation or to small resolution reduction \cite{srivastava2019minimal}.
Our work relates to the psychophysical study mentioned above \cite{ullman2016atoms} as we use our model to both evaluate the MIRCs and to suggest computationally specified locally minimal recognizable patches, denoted cMIRCs, which exhibit similar properties to the human-specified MIRCs.

The amount of information required for algorithmic recognition was considered in several works. Image regions of intermediate complexity were found to be maximally informative \cite{ullman2002visual}. Sensitivity of common CNN architectures to occlusions was evaluated in various studies  \cite{pepik2015holding,geirhos2018generalisation,tang2018recurrent}. 
Robustness to different image distortions such as blur, noise, contrast and JPEG compression was also examined \cite{dodge2016understanding,geirhos2018generalisation}. It seems that local features carry a lot of information and suffice for high accuracy classification  \cite{brendel2019approximating}.

\begin{figure*}[t]
	\centering
	\includegraphics[width = 0.9\textwidth]{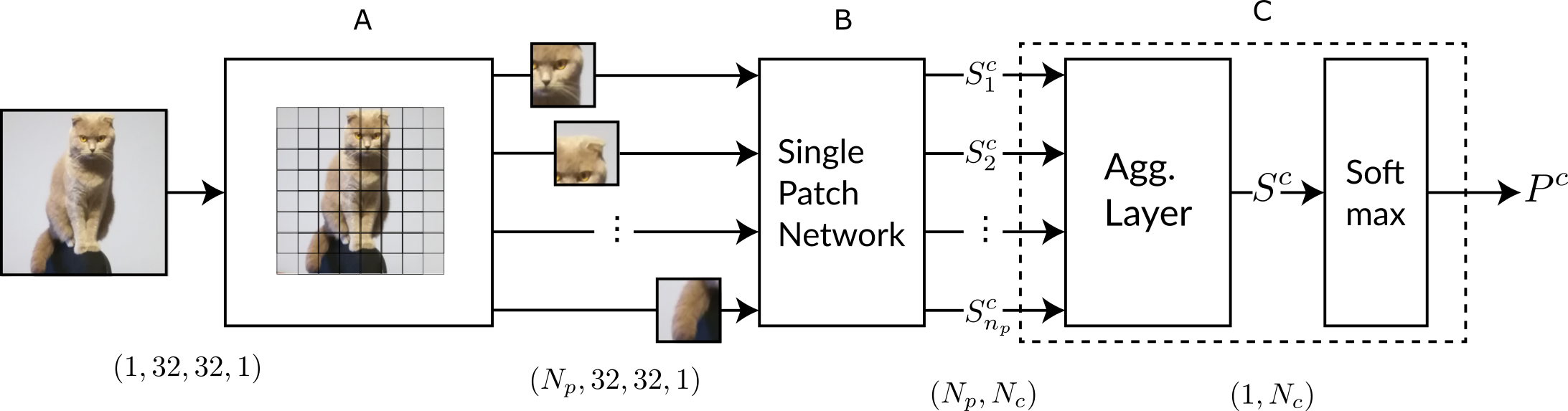}
	\caption{PBC model - (A) Input image is split into patches which are resized to a standardized size. (B) Each patch is analysed by the SPN. (C) The aggregation layer transforms patch-level scores into image-level scores and a softmax layer converts them into image-level class-probabilities.}
	\label{fig:PBC Model}
\end{figure*}

\section{Tools for Determining Patch Recognizability} \label{sec:Recognizability} 

\subsection{Patch Recognizability} \label{sec:Recognizability Definition}
Our goal is to characterize the globally and locally minimal sub-image (patch) required for successful categorization. 
In this paper, we consider this general question in the context of a specific data set and in the closed set setting.

To address image variation, due to scaling and other pose changes, we specify patch size as a fraction of the full object size, as seen in the given image. 
In our study, we use fixed size images ($32 \times 32$) from the CIFAR datasets. Most images contain one object, tightly bounded by the image boundaries. For these images, a fixed fraction of the object size corresponds to a fixed size in pixels.

Consider a specific patch. We shall say that this patch is {\em Locally Recognizable} if a categorization procedure accepting only this patch as its input  classifies it to the correct category. 
Formally, let $S_p^c$ denote the score of class $c$ associated with the patch $p$. Here this score is provided by a CNN, denoted a single-patch-network, described below. Then the patch is {\em Locally Recognizable} relative to this score if the inferred class, 

\begin{equation}
	{\hat c}=argmax_c \  S_p^c
	\label{eq:Locally Recognizable Patch}
\end{equation}
is correct. 

Note that this notion of recognizability is rather weak. A patch can be small or smooth and get similar scores for several categories. Yet, if the score associated with the correct category is a bit larger than the other, it is, by this definition, recognizable. The following definition is more meaningful: A 
patch is {\em q-Locally Recognizable} if a categorization procedure accepting this patch as its input  classifies it to the correct category with a confidence larger than {\em q}. We shall use confidence measures defined below. 

We are also interested in global, image-level, recognizability from an image patch. 
A sub-image is {\em globally recognizable} if the correct class score associated with this patch is higher than all other scores associated with all other patches and with other classes. Thus, when the image contains at least one globally recognizable patch, then the estimates class
\begin{equation}
	{\hat c}=argmax_c \ (max_p \  S_p^c)
	\label{eq:Globally Recognizable Patch}
\end{equation}
is correct. 

In contrast to local recognizability, it is unlikely that a very small or smooth patch would be associated with the maximum global score. This is because typically, numerous similar patches will be present in images of other categories. On the other hand, for categories that are not too similar, we expect to find in each image a sub-image of sufficient size and detail that is consistent only with its category. 

\subsection{The Patch Based Classification Model} \label{sec:PBC Model}
The main computational tool developed for the study of minimal recognizable patches is the patch-based classification (PBC) model, which calculates scores and confidences, and performs image-classification based on information included in the best or most informative single patch of the full image. 
The best patch is unknown and is not pre-specified; therefore, locating it is part of the network's task. This also means that learning the PBC classifier is a weakly supervised task.
The model is aimed at finding the globally minimal patch. Yet, its learning process provides scores and confidences that are used for finding locally minimal patches as well. Aiming at global recognizability is a harder task, and as we found, training for it also produces better classifiers for local recognizability.

We use different networks, sharing a fixed architecture, for each patch size.  The network is composed of the following parts: a. Splitting the input image into $N_p$  (overlapping) spatial patches and resizing each one to a standardized size. b. Independently analyzing each patch using a CNN, denoted the single-patch-network (SPN). For each patch, the SPN provides $N_c$ patch-level scores, one for each category. c. An aggregation layer converting the patch-level scores of all $N_p$ patches into $N_c$ image-level scores, which are normalized by a softmax layer, providing $N_c$ image-level class-probabilities. See Fig. \ref{fig:PBC Model} for a diagram of the model. We elaborate on these network parts below. 

\subsection{Single Patch Network} \label{sec:Patch-Network}
The patch network could be any standard classification network, and finding the most accurate network is not a goal of this paper. The network used in our experiments follows the All-Convolutional-Net model \cite{springenberg2014striving}.
Relatively simple in nature, this model achieves high accuracy  just a little lower than the best much more complex  classifiers. We modified it slightly by replacing the dropout regularization layers with batch-normalization and the $6\times6$ global-averaging layer with a more generalized $6\times6$ convolutional layer. 
The softmax layer was moved out of the single-patch-network, to be placed after the aggregation stage (see Section \ref{sec:Softmax Location}).

Comparing responses and accuracies for different patch sizes is essential in this study. Therefore, to avoid an architecture-dependent bias, we insisted on using a uniform architecture (with different learned weights) and on interpolating the patches to the same input size: $32 \times 32$. As expected, when experimenting with other interpolated input sizes, we found that smaller interpolated patches work somewhat better with smaller original patch sizes, for which the interpolation is less extreme. However, the small differences in the accuracy (less then $5\%$) were not significant for this study.

\subsection{Patch Score Aggregation} \label{sec:Aggregation Methods}
The scores for all patches and all categories (patch-level scores) are aggregated to give image level scores, one for each category.  
We considered two types of max score aggregation:

\begin{description}
	\item [Category-independent max - ] ($\boldsymbol{S}_{max-ind}^{c}$) This score, evaluated separately for each category, is the maximal score of this category over all patches.  For this aggregation, the score for each class is usually taken from a different patch. 
	\item [Winner-directed max - ] ($\boldsymbol{S}_{max-dir}^{c}$)  The image score for all classes is taken from a single patch, the one associated with the overall maximum score.
\end{description}

Both aggregation methods classify the image using the best overall score, as specified in Equation (\ref{eq:Globally Recognizable Patch}).
The first uses patches that are possibly different from the winner, for evaluating the scores associated with other classes.
The second aggregation takes the other classes' scores from the same patch, ignoring possibly higher scores from other patches.  
Formally, 
\begin{align}
	S_{max-ind}^{c} &= \max_p{\{S_{p}^{c}}\}, \\
	S_{max-dir}^{c} &= S_{p^*}^{c}\ , {\rm where} \ 
	p^* = argmax_p \ (max_c \ S_{p}^{c}) 
	\label{eq:Aggregation}
\end{align}

The aggregation method influences the confidence associated with the different categories, prediction loss, and hence the training process. 

It seems that an intelligent agent wishing to categorize the object(s) in a scene, would scan that scene and try to extract the best evidence for each category, no matter where. In this context, calculating confidence using the first, category-independent aggregation, is justified. On the other hand, when only one patch is observed, the second, winner-directed, aggregation describes the available information better. Moreover, in scenes containing several objects, the second aggregation allows the detection of multiple categories. For such scenes, it also helps the learning process, because the presence of an object from one category on one place does not indicates that responses to other categories in other locations should be suppressed. Empirically, the two aggregation methods give similar results, with some advantage to the first. See Section \ref{sec:Agg-Effect} for more details.

\subsection{Placing the Softmax Layer} \label{sec:Softmax Location}
The softmax score normalization is the final layer, acting on the image-level scores provided by the aggregation layer.
In principle, we could alternatively apply softmax normalization on the scores of every patch separately, before aggregation. This however, would let the response to other classes influence the maximum. A substantial, but not maximal response to some class, for example, would lower the normalized response to the winning class, which might otherwise be larger than the response to this class in all other patches. 
We also found experimentally that using a local, patch-level softmax layer hurts the model's generalization ability.

\section{Experimental Setup} \label{sec:Experimental}

\subsection{The CIFAR$10^*$ Dataset} \label{sec:Custom Dataset}
We started our experiments with the CIFAR10 dataset, containing ten classes \cite{krizhevsky2009learning}. 
Eight of these classes can be divided into pairs of related and similar categories: ship-plane, car-truck, dog-cat and horse-deer.
We observed that for small patches, the learned model often preferred one of two similar categories and “gave up” on the second one. 
It seems that informative small patches of related categories (e.g, the wheels in automobiles and trucks) were effectively indistinguishable for the classifier.
By choosing the class with larger amount, or clearer, appearance of this patch, the classifier  achieves better mean performance. This observation points at a limitation of recognizing from a patch.
This phenomenon interferes with finding minimal recognizable patches for the non-preferred categories. Therefore, we experimented with a CIFAR10 variant, which is easier in the sense of containing less inner-similarities. 
This variant, denoted CIFAR$10^*$, was based on classes from the CIFAR10 and CIFAR100 datasets, and consisted of $3,000$ samples from each of the following classes: airplane, automobile, bird, cat, deer, frog, fish, tree, person and insect. The data set was divided into a training set of $25,000$ images and a test set of $5,000$ images, both well-balanced between the $10$ classes.

\subsection{Training and Implementation} \label{sec:Training}
We used a grayscale version of the CIFAR$10^*$ dataset to train $16$ patch-based models with the first aggregation (category-independent max) for $16$ square patch sizes, $d_i \times d_i$ pixels, where $d_1=32, d_2=30, \dots, d_{15}=4, d_{16}=2$.  We refer to the models simply as "model of size $d$". The model of size $32$ corresponds to the full image.
These grayscale input, category-independent aggregation-trained models are the default models used throughout our experiments.
We also trained models with color inputs and using the (second) winner-directed aggregation. 

During training, the spatial stride taken while splitting each image was set to be half the patch size, except the smallest, $2\times2$ patch, for which a $2\times2$ stride was used.
For evaluation, the stride was always set to be a single pixel.

For training, we used a categorical cross-entropy loss function with an Adam optimizer. Training was conducted for $150$ epochs, with an initial learning rate of $0.001$, reduced by a factor of $2$ every $30$ epochs. Weights were regulated by ridge regression with a $1e-4$ coefficient. The batch size was set to $50$ images.

The same hyper-parameters were used for the training of all models. We found that models working with smaller patches can benefit from lower regularization. However, the difference was small (less then $4\%$) and not significant for this study. 

\subsection{Preliminary Results} \label{sec:Perlim Results}
\begin{figure}[b]
	\centering
	\includegraphics[width = \columnwidth]{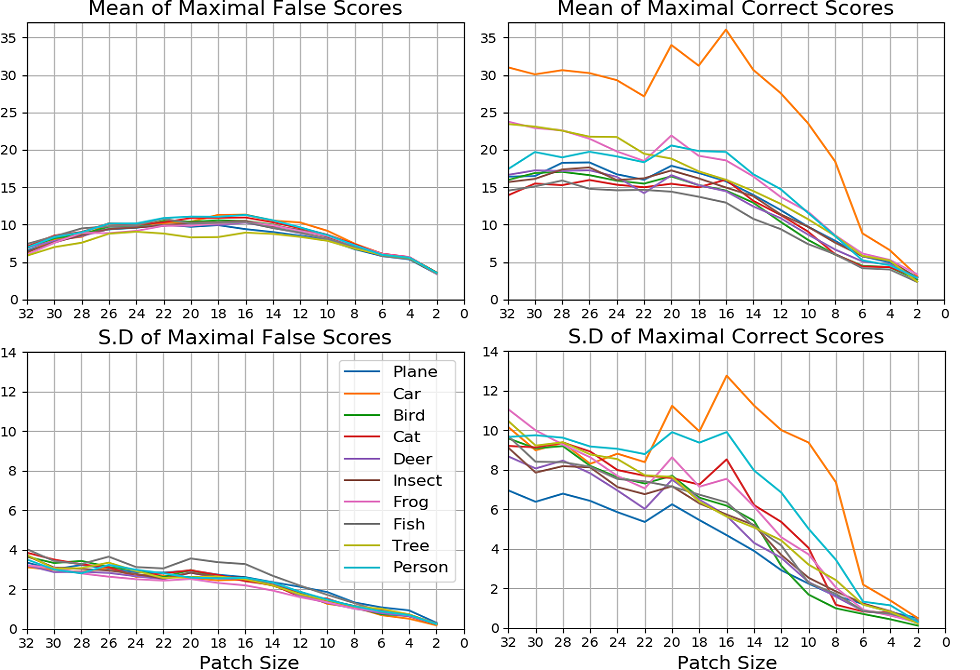}
	\caption{Mean and standard deviation of maximal false scores (left) and maximal correct scores (right) as functions of patch size.}
	\label{fig: score_mean_std}
\end{figure}

We evaluated the pre-softmax output of the PBC model, associated with the correct and incorrect category classifiers. As expected, correct category classifier scores are higher. 
Interestingly, the maximal scores associated with incorrect classifiers (with category-independent aggregation) are relatively consistent among different categories and patch sizes (see Fig. \ref{fig: score_mean_std} (left)).  In particular, small patches get almost constant small scores, as apparent from the low standard deviation.
This observation is used later in the cMIRC specification process.
The distribution of all false scores is much wider.

\section{Patch Size and Recognizability} \label{sec:Exp-Recognizable Patches}

\subsection{Single-Image Recognizability} \label{sec:Exp-Confidence}

\begin{figure*}[t]
	\centering
	\includegraphics[width = \textwidth]{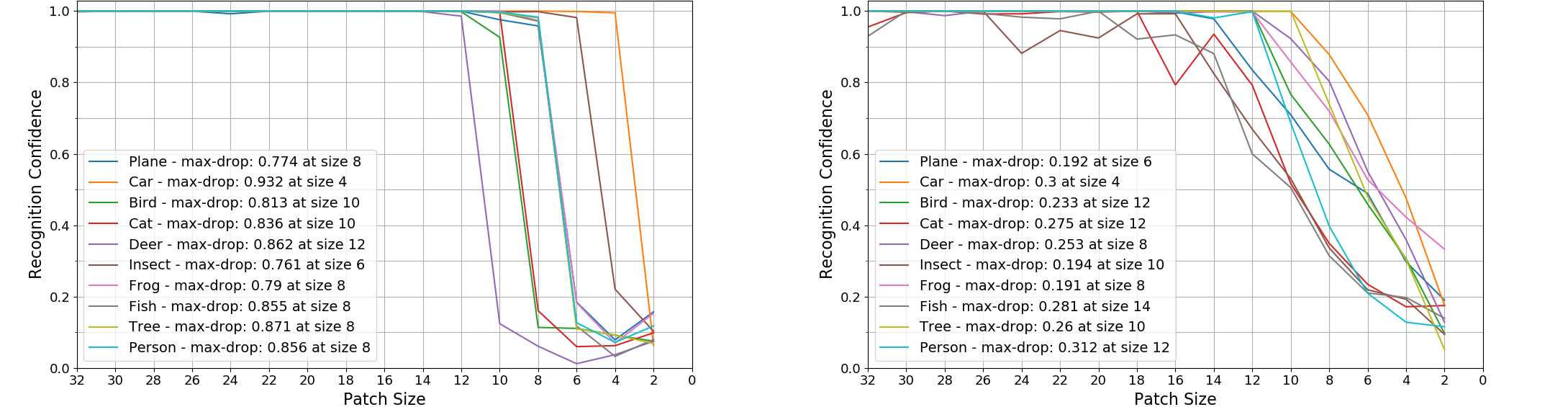}
	\caption{Confidence curves of images with the largest (left) and smallest (right) maximal drops.}
	\label{fig:Confidence Exp}
\end{figure*}

Ideally, we would like to estimate the recognition accuracy as a function of patch size. For a particular image, the accuracy cannot be estimated empirically and is substituted by a single image accuracy estimate, or confidence. In networks trained with cross entropy loss, the softmax response approximates the posterior probability for this category and may be used as a simple and yet reasonably accurate confidence \cite{geifman2018bias}.  
We evaluated correct class confidence as a function of patch size and further measured the confidence  difference between two consecutive patch sizes $d_{i} \longrightarrow d_{i+1}$. 

The confidence curves for specific images reveal sharp, significant confidence drops in most images. For each image, we refer to the maximal confidence drop associated with two consecutive patch sizes, and denote this shortly as the maximal drop.
The curves describing the confidence in the images associated with the largest and the smallest maximal drop (two images for each category) are plotted in Fig. \ref{fig:Confidence Exp}. 
A small part of the images were associated with a smooth uniform confidence decrease and small maximal confidence drop (see Fig. \ref{fig:Confidence Exp} (right)). Some other images were difficult to classify even as full images, and were associated with low, smooth confidence curves. 

\begin{figure}[hb]
	\centering
	\includegraphics[width = \columnwidth]{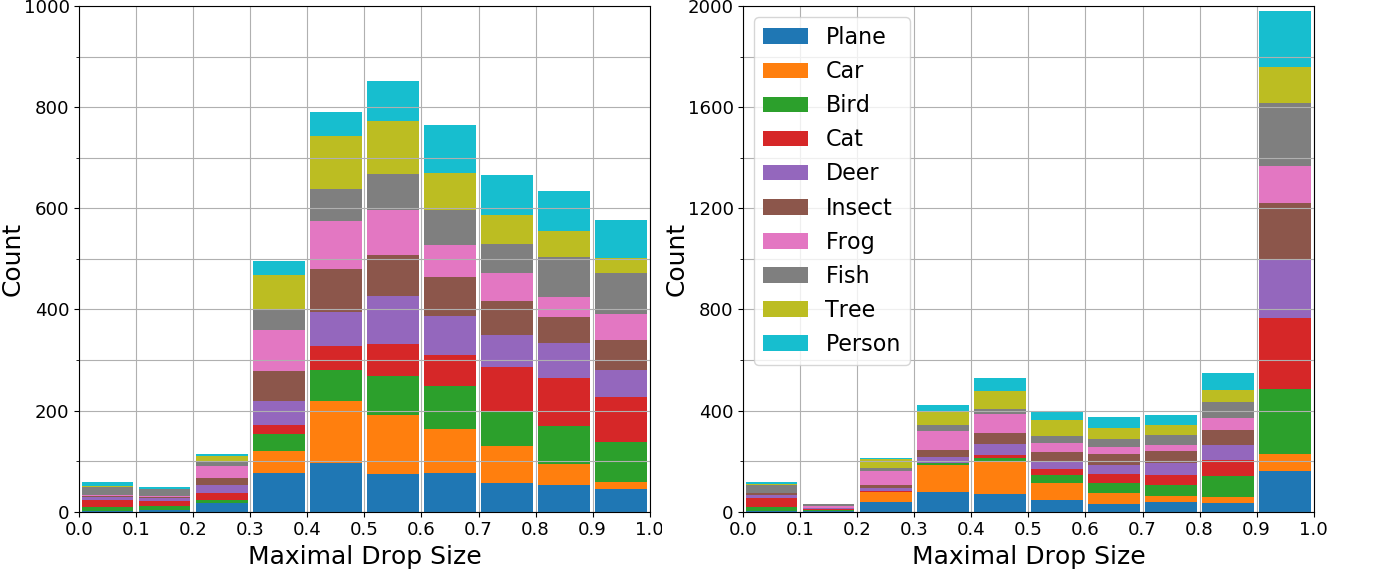}
	\caption{Histograms of maximal confidence drops for classifiers trained with the category-independent (left) and winner-directed (right) aggregations.}
	\label{fig:Drop-Hist 1D}
\end{figure}

For most images, however, the maximal confidence drop is substantial, as revealed in the maximal drop histogram (see Fig. \ref{fig:Drop-Hist 1D}). Specifically, for the majority of images ($3,492$ out of $5,000$), the maximal drop was larger than $0.5$. 

\begin{figure}[b]
	\centering
	\includegraphics[width = \columnwidth]{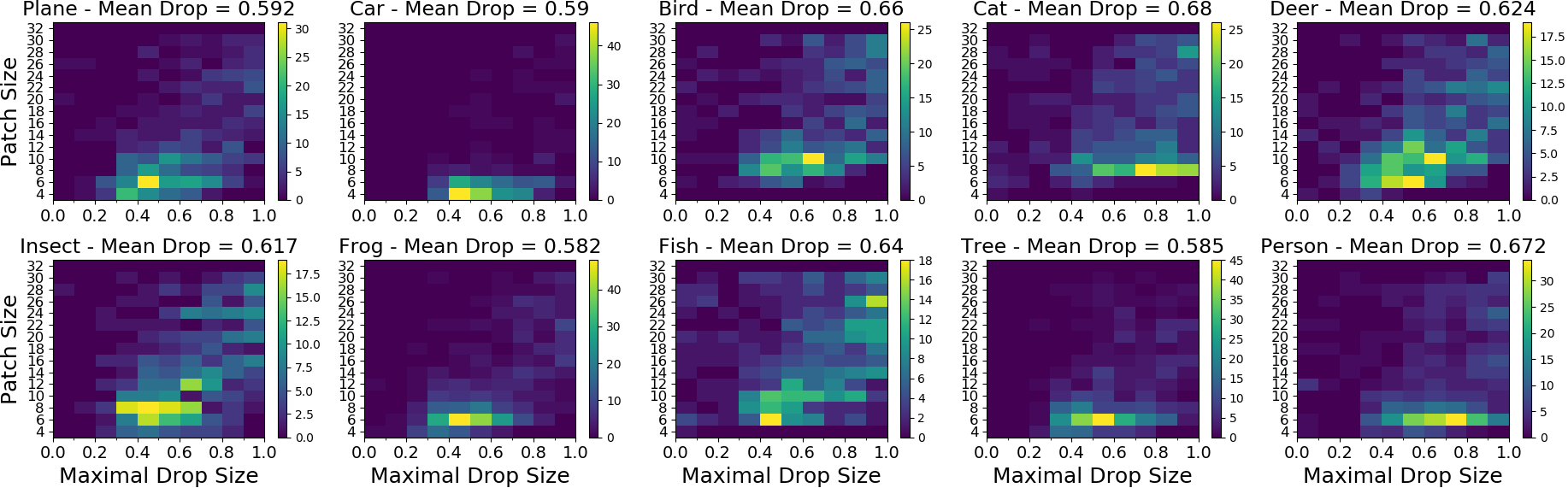}
	\caption{2D histograms of maximal confidence drop (x-axis) and its associated patch size (y-axis), for classifiers trained with the category-independent aggregation.}
	\label{fig:Drop-Hist 2D}
\end{figure}

The critical patch sizes associated with the maximal confidence drop are not fixed. Even images of objects from the same category differ a lot due to the intra-class variability and the uncontrolled object pose. 
To show this variation, we plotted 2D histograms of the maximal drop size and the corresponding (larger) patch size $d_i$  (see Fig. \ref{fig:Drop-Hist 2D}). Clearly, the size of this critical patch varies significantly over the set of images associated with each category, and for some categories there are even several dominant sizes. The images' maximal drops varied as well, and were typically larger when they occurred with larger patches.

This behavior was reproduced for both grayscale and color images. The average maximal drop was slightly different: $0.608$ for color vs. $0.624$ for grayscale, and patch size associated with the maximal confidence drop was smaller for color. That is to say, as expected, larger patch sizes are required for recognition from gray level images \cite{torralba200880}.

The classifier trained with the second, winner-directed, aggregation led to even more substantial maximal drops: $0.72$ on average. The number of images with a maximal drop larger than $0.5$ increased as well ($3,683$ vs. $3,492$).
This is not surprising, because for each incorrect category,  the score associated with the patch corresponding to the winner is, by definition, lower than the score obtained for the best patch in this category. 
This makes the confidence higher. The confidence drop tends also to be higher because it is a difference between two winner-directed confidences, each of which is higher than the corresponding category-independent confidence.

\subsection{Category Recognizability} \label{sec:Exp-Categorization}

\begin{figure}[hb]
	\centering
	\includegraphics[width = 0.95\columnwidth]{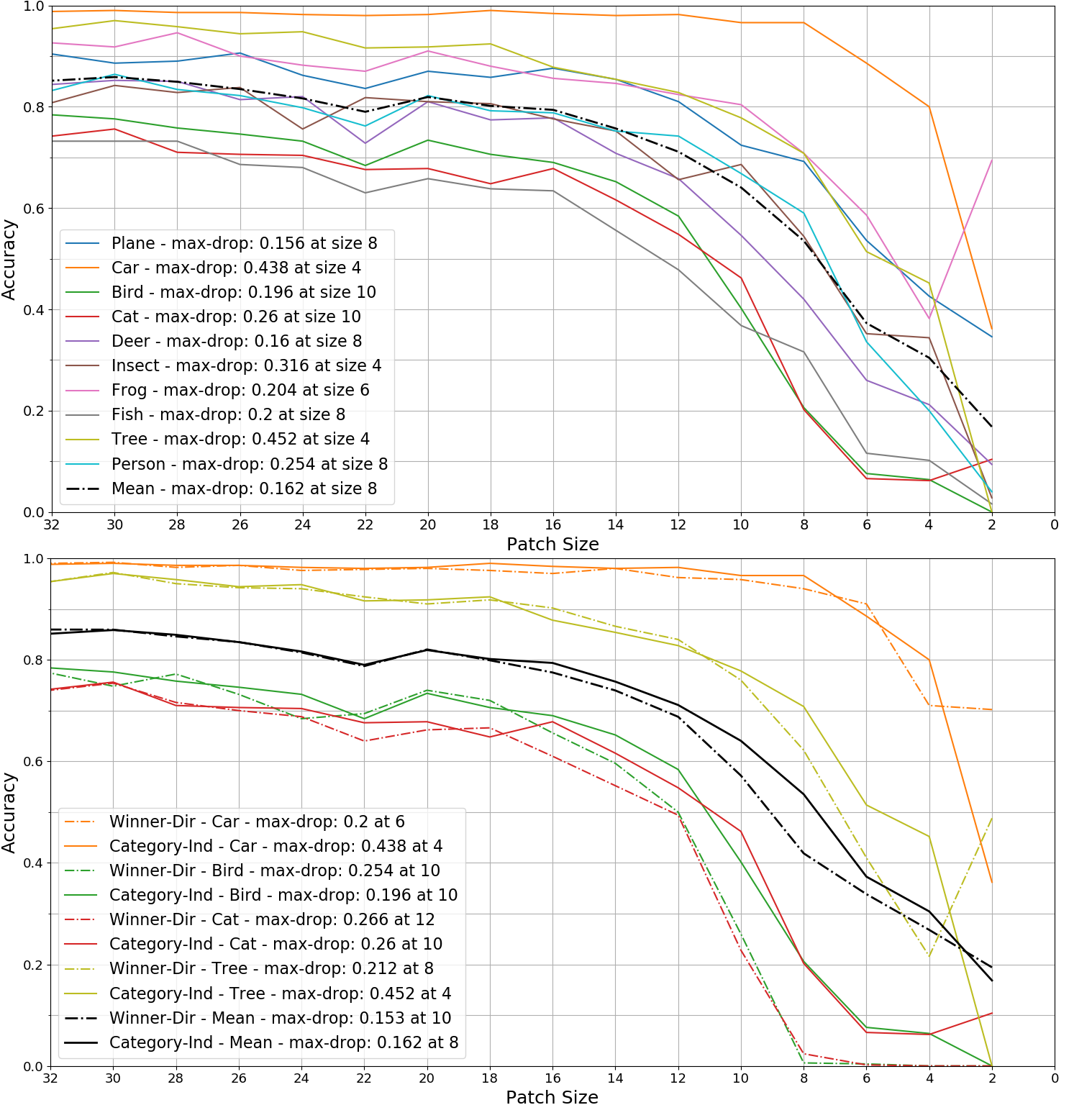}
	\caption{Mean and class-specific categorization accuracy with category-independent aggregation (top), and its comparison with winner-directed aggregation (bottom).}
	\label{fig:Categorization Exp}
\end{figure}

We carried out additional experiments designed to evaluate the accuracy of categorization from a single patch. 
This accuracy is estimated simply as the  fraction of images for which the PBC model provides the correct label.  
As expected, smaller patches provide less information and lower accuracy (see Fig. \ref{fig:Categorization Exp}).
Remarkably, all categories were classified correctly with $50\%$ accuracy with $12\times12$ patches, corresponding to roughly $0.14$ of the image area. 

Considerable variation exists between the curves of different categories as well. Some categories may be identified from very small patch sizes, which may correspond to either distinct small features (e.g. a wheel or an eye) or to texture (tree foliage). Other categories required a coarser scale structure (e.g. birds and cats).

The variability of critical patch sizes within each category, observed in the histograms, imply also that different images of the same category may need different patch sizes to be  recognized reliably. The fraction of images associated with a sufficiently large patch grows slowly with the patch size, and corresponds to the smooth accuracy curves seen in Fig. \ref{fig:Categorization Exp} (top). 
The critical patch size variability (within category) is influenced by the variance of the available instances. Categories with low  appearances variance (e.g. cars, which are typically photographed from specific viewpoints) showed higher accuracy over all patch sizes, and in particular, benefit also from small discriminative features. 

As expected , similar tests with color images, produced higher accuracy and weaker dependency on patch size.
See \cite{torralba200880} for a study revealing the advantage of color in low-resolution images. 

\subsection{The Aggregation Effect on Categorization Accuracy} \label{sec:Agg-Effect}
An important part of the proposed patch-based model is the aggregation, which can be either category-independent, or winner-directed (see Section \ref{sec:Aggregation Methods}).
The choice of the aggregation method influences the resulting classification accuracy. Note that the accuracy difference is only due to different training, because once the classifier is trained, the aggregation provides the same winner: the class associated with the highest score in any of the patches (Equation \ref{eq:Globally Recognizable Patch}).

The differences in results between the two aggregations are small (see Fig. \ref{fig:Categorization Exp} (bottom)).
The small difference is still noteworthy, however, because in training with category-independent aggregation, the best patches for each incorrect class are used for suppressing the score to this class. Using these patches and not the particular winner-directed patch is much more informative and leads to better SGD steps, more stable training, and faster convergence. Yet, with the less informative winner-directed patches, the obtained accuracy is almost as good.
It turns out that with smaller patches, the overlap between these best patches and the winner patch is smaller, the difference is potentially larger, and the disadvantage of learning with winner-dependent aggregation is more significant. The difference in accuracy is correspondingly larger, but it is still small.

\section{Minimal Recognizable Image-Patches} \label{sec:Exp-Minimal Patches}

\subsection{Globally Minimal Recognizable Patches} \label{sec:Exp-MRP}

\begin{figure}[t]
	\centering
	\includegraphics[width = 0.8\columnwidth]{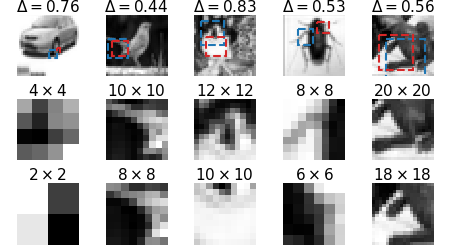}
	\caption{MRP examples. Top: Images with their MRP (blue), best smaller unrecognizable patches (red), and the associated confidence drop. Middle: MRP sub-image. Bottom: best unrecognizable sub-image.}
	\label{fig:MRP Example}
\end{figure}

For all images, the correct classification confidence decreases with the patch size. Let $d^*$ be the minimal patch size for which the image classification is correct. (Such a minimal patch exists for almost all images, with the exception of a few images that are not classified correctly even from a full image).
With some abuse of notation, we denote one of these recognizable patches (of size $d^*$) -- the one associated with maximal score (and confidence) -- as the minimal recognizable patch (MRP).
By definition, the MRP is unique and of globally minimal size. Other globally recognizable image patches of the same critical size $d^*$, but with somewhat lower scores, are often present in the image.
As described in Section \ref{sec:Exp-Confidence}, for a majority of images, there is a sharp confidence drop larger than $0.5$. Let $(d_i, d_{i+1})$ be the pair of patch sizes associated with this maximal drop. 
Interestingly, while the MRP is determined exclusively based on recognizability, for most images ($3,418$ out of $5,000$) it is associated with the maximal confidence drop; that is $d^*=d_i$.
Several MRPs are shown in Fig. \ref{fig:MRP Example}. Some of them are consistent with human judgment which can identify the category from the MRP but not from the best smaller patch. For most MRPs, however, this consistency is weaker.

\subsection{Locally Minimal Patches} \label{sec:Exp-MIRC}
An important motivation for our study was the sharp accuracy drops associated with decreasing patch sizes observed in human vision experiments, but not with recognition  algorithms \cite{ullman2016atoms}.
In our single image confidence experiments, however, we found sharp confidence drops, but these experiments consider globally minimal patches and are therefore different than the MIRCs considered in both perceptual and computational tests in \cite{ullman2016atoms}.

\subsubsection{Computational MIRCs}
To get closer to the tests performed in \cite{ullman2016atoms}, we now consider local, MIRC like patches. Our goal here is to specify them, find their distribution in the image, and test whether sharp confidence drops arise. 

We focus on patches associated with correct classification. Thus, we start by using the learned single-patch-network (SPN) to calculate correct class confidence for every image patch (of every size). 
It turns out that calculating confidence for every patch is not straightforward. This is because most patches, and especially the smaller ones, are non-informative, while the SPN was trained mostly on the most informative patches in the image. Therefore, the SPN gives arbitrary scores for non-informative small patches and the resulting softmax confidence may be occasionally erroneously high, leading to the false conclusion that the patch is informative. 
Note that even if the classifier would give a low score for all classes, it could be that the softmax ratio would be still high. 
Thus we take an indirect approach and estimate the confidence as follows: for calculating the correct class confidence of a patch, we take the score (SPN output) from this patch. However, for the incorrect classes, we set the scores as the maximal scores associated with possibly other patches in the image. This is similar to the category-independent (global) aggregation used in the training process. Here, however, as we do not have access to the rest of the image, we substitute these scores with their expected values. As we saw in Section \ref{sec:Perlim Results}, these values are roughly constant, and do not depend on the particular image chosen. 

We argue that the usage of learned typical scores is legitimate. It makes the decision closer to open set classification and is more consistent with the tests performed in \cite{ullman2016atoms}. It is likely that humans,  attempting to recognize an object from a partially visible object, use past experience for calculating their confidence and their final decision. 
Following the MIRC definition \cite{ullman2016atoms}, a cMIRC patch is specified as one that is q-locally recognizable (see Section \ref{sec:Recognizability Definition}), while all its nine contained sub-patches (obtained by cropping each spatial dimension by two pixels) are not. We experimented with several $q\in [0.2,0.7]$ values and got similar results. The following results correspond to  $q=0.5$.

\begin{figure}[b]
	\centering
	\includegraphics[width = 0.95\columnwidth]{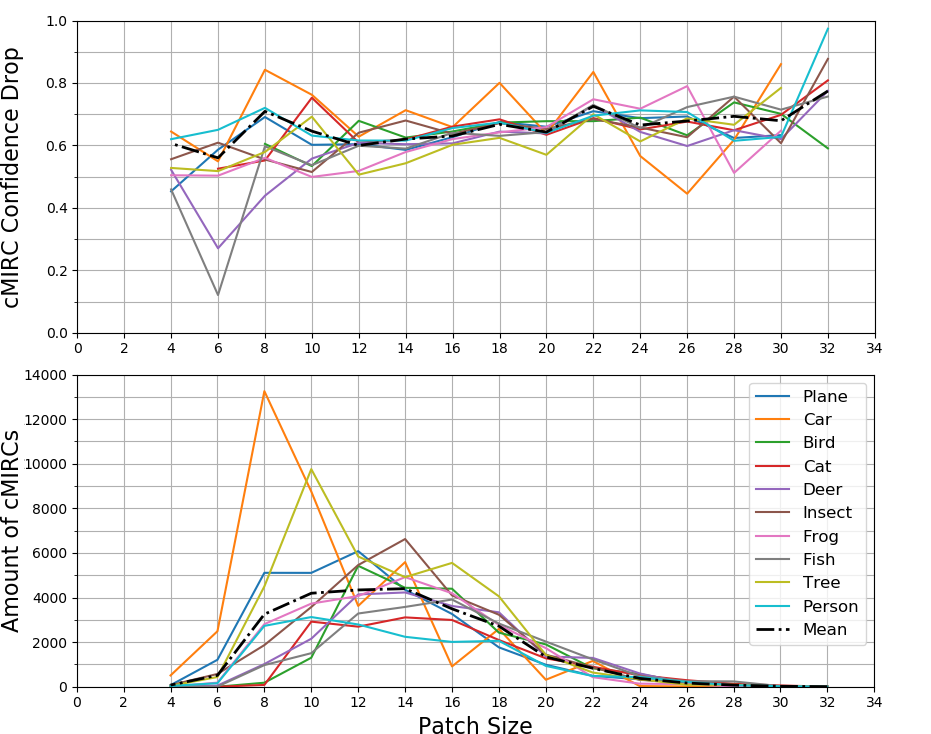}
	\caption{Mean drop in confidence between a cMIRC and its best (out of nine) contained sub-patch (top) and the amount of discovered cMIRCs (bottom) as functions of a cMIRC's patch size.}
	\label{fig: cMIRC}
\end{figure}

We found that each image from the CIFAR$10^*$ test dataset contains multiple cMIRCs of different sizes and positions. On average, we found $51.5$ cMIRCs per image.
This number is bigger than the $15.1\pm7.6$ MIRCs per image found in \cite{ullman2016atoms}. 
However, if we exclude cMIRCs with shifts of a single pixel, we revert to $20.6$ cMIRCs per image (comparable with \cite{ullman2016atoms}).
As expected, we found cMIRCs of several sizes (see Fig. \ref{fig: cMIRC}), and one of the smallest cMIRCs typically coincides with the MRP. 
The average confidence drop between each cMIRC and its best sub-patch (over all images) is $0.64$.

These results are similar to those observed in the human vision study \cite{ullman2016atoms}, and suggest a simplistic, feedforward model to the perceptual mechanism. 
An explanation why previous algorithmic attempts did not reproduce the sharp accuracy drop is discussed below.   

\subsubsection{Evaluating MIRCs with the PBC Model}
Typically, MRPs and some of the cMIRCs are relatively small and difficult for humans to recognize (see Fig. \ref{fig:MRP Example}). The MIRCs specified in \cite{ullman2016atoms} are more recognizable. We hypothesize that the reason for this difference is the easier tasks considered in our computational study, where close set classification with only 10 categories were considered. In the human study, however, the task was to recognize objects from an unlimited library, which is harder. Therefore, more informative patches, of larger size and detail, were specified as MIRCs. This probably implies that their sub-MIRCs are also recognizable by computational closed set algorithms, which, in turn, imply that the accuracy change, between MIRC and sub-MIRCs, is not significant. 

To test this hypothesis, we took two images from \cite{ullman2016atoms} that belong to the CIFAR$10^*$ categories: a airplane and a bird, and tested all their MIRCs. 
The MIRC sub-images were estimated manually from the original paper.
Each MIRC was resized to a size of $32\times32$, and split into $49$ sub-MIRC patches (each of size $26\times26$ ($81\%$)). The correct class confidences and the confidence drop between the MIRC and its best sub-MIRC patch were evaluated using our PBC model.
Since MIRCs are originally of different sizes (Section \ref{sec:Exp-Categorization}), this was repeated for all PBCs, corresponding to all input sizes and the largest drop was kept.
The mean and maximal confidence drop for all MIRCs were $0.11$ and  $0.3$, respectively. These drops are considerably smaller than drops associated with MRPs and cMIRCs, but are consistent with the finding in \cite{ullman2016atoms}.

\begin{figure}[t]
	\centering
	\includegraphics[width = 0.9\columnwidth]{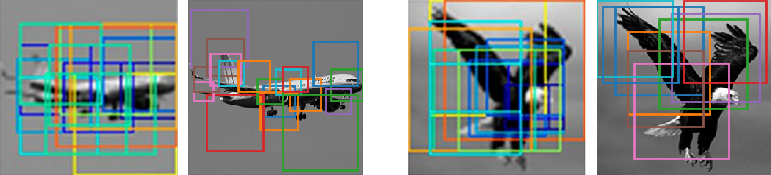}
	\caption{Two images from \cite{ullman2016atoms} with their human-specified MIRCs (left) and computationally specified cMIRCs (right).}
	\label{fig:Human MIRC}
\end{figure}

\section{Conclusions} \label{sec:Conclusions}
This work empirically characterizes the globally minimal sub-image required to categorize an image successfully. A specialized deep network that learns by a weakly supervised auxiliary task was designed for this task.
We show that the size this minimal sub-image takes, on average, is a small fraction of its full area, but also that it varies significantly within each category. Following a human vision study \cite{ullman2016atoms}, another type of minimal recognizable patches that are not globally minimal, but are (locally) minimal in the sense that no sub-patch of them  is recognizable, was specified as well.  

Both types of minimal recognizable patches share a surprising common property with the human vision study described in \cite{ullman2016atoms}: there are image regions that are sufficiently informative for recognition, but which stop providing the required information for recognition following a small size decrease. Moreover, the reduction in region informativeness is sharp and substantial.
Remarkably, in both studies, this sharp reduction was not part of the demands but was found, empirically, as a byproduct. 
Earlier work did not succeed to computationally reproduce the perceptual sharp reduction effect \cite{ullman2016atoms} (see however \cite{srivastava2019minimal}). 

In contrast to previous work, which provide specific  informative image patches (e.g., the  "fragments" in \cite{vidal2003object}) characterizing the images of certain categories, we provide here a generalized and somewhat different characterization: a network that gives a high, discriminative score to the informative patches. Thus the informative patches themselves are only implicitly characterized, and the best image patches (of the same size and category) may be   different in different images.

The minimal recognizable patches we found were small, and  usually  unrecognizable by humans. This is due, in our opinion, to the closed-set setting and  the small number of classes. We intend to estimate MRPs that are more consistent with human vision using more classes or open-set classification tools.


\bibliographystyle{IEEEtran}
\bibliography{article_bib}

\end{document}